# Predicting Penalty Kick Direction Using Multi-Modal Deep Learning with Pose-Guided Attention

Pasindu Ranasinghe[1] and Pamudu Ranasinghe[2,3]

[1] University of New South Wales, Sydney, Australia
pasindu.ranasinghe@unsw.edu.au
[2] Virtusa Pvt. Ltd., Colombo, Sri Lanka
[3] University of Moratuwa, Moratuwa, Sri Lanka

**Abstract.** Penalty kicks often decide championships, yet goalkeepers are left to anticipate the kicker's intent from subtle biomechanical cues unfolding within a narrow time window. This study presents a real-time, multi-modal deep learning framework for predicting the direction of a penalty kick—left, middle, or right—prior to ball contact. The model adopts a dual-branch architecture: MobileNetV2-based CNN extracts spatial features from RGB frames, while 2D keypoints are processed using an LSTM network with attention mechanisms. Pose-derived keypoints are further used to guide visual focus toward task-relevant regions. A distance-based thresholding method segments input sequences immediately before ball contact, providing consistent input across diverse footage. A custom dataset of 755 penalty kick events was curated from actual match videos, with frame-level annotations for object detection, penalty shooter keypoints, and the final ball placement in the goal. The model achieved 89% accuracy on a held-out test set, outperforming visual-only and pose-only baselines by 14–22%. With an inference time of 22 milliseconds, the lightweight and interpretable design makes the model well-suited for goalkeeper training, tactical analysis, and real-time game analytics.

**Keywords:** Penalty kick prediction, sports analytics, multi-modal learning, computer vision, action recognition

## 1 Introduction

Penalty kicks are among the most critical moments in football, often shifting the momentum or deciding match outcomes in high-stakes competitions. The success of a penalty attempt traditionally depends on the kicker's skill and strategy versus the goalkeeper's reflexes and anticipation. However, the inherent variability of human performance makes the outcome of any given penalty kick difficult to predict.

Recent advances in artificial intelligence and computer vision have opened new frontiers in sports analytics, enabling data-driven decoding of complex athletic movements [1]. Deep learning-based computer vision techniques can analyse video footage frame by frame to extract subtle cues invisible to the naked eye [2]. Despite these advances, predicting the direction of a penalty kick from video remains an open challenge. This requires interpreting complex spatial dynamics—from run-up angle and body posture to foot positioning and hip rotation, all of which can influence the outcome [3].

To address these challenges, this paper proposes a multi-modal deep learning framework that integrates both visual context and the pose dynamics of the penalty taker. The architecture combines spatial features extracted from RGB frames using a MobileNetV2-based CNN, while an LSTM network models the temporal evolution of 2D body keypoints. A pose-guided spatial attention mechanism further sharpens focus on relevant player actions and scene-level cues. By fusing scene-level and biomechanical data, the model predicts the direction of a penalty kick—left, middle, or right—with an accuracy of 89%.

## 2      Related Work

Recent developments in football analytics increasingly rely on data-driven methods to support player evaluation, outcome prediction, and tactical decision-making [1, 3, 4]. Among these, deep learning techniques have shown significant promise in automating the visual interpretation of match footage. Convolutional Neural Networks (CNNs) are commonly employed for detecting and tracking players and the ball, while Recurrent Neural Networks (RNNs), such as Long Short-Term Memory (LSTM) networks, are used to model temporal patterns in play sequences [5-7]. For instance, Honda et al. demonstrated that combining visual features with player trajectories can significantly enhance the prediction of pass receivers in soccer [7].

In the specific context of penalty kick analysis, Chakraborty et al. [8] proposed a YOLOv4-based object detection pipeline combined with OpenCV tracking, followed by LSTM models to analyse body positioning data during penalties. Their approach achieved a mean accuracy of 79.05% one second before the kick [8]. More recently, Salazar and Alatrista-Salas introduced a dedicated penalty kick dataset and used semantic segmentation along with 3D pose estimation to train deep models for shot placement prediction [9]. Pinheiro et al. integrated body pose estimation using OpenPose to detect relevant body orientation angles that correlated with goalkeeper anticipation [3]. While prior research has shown encouraging results in penalty kick analysis, several important challenges remain unresolved. A common limitation is that many existing models have been evaluated only in offline conditions, reducing their effectiveness in real-time match scenarios. Although some studies incorporate pose estimation, this is often treated as an isolated component, rather than being deeply integrated into the prediction framework. This lack of integration limits the model's ability to capture fine-grained biomechanical cues that arise from the interaction between body posture and scene context. Evidence from sports biomechanics literature highlights that features such as trunk orientation and kicking-foot height are strong predictors of shot direction, reinforcing the need for pose-informed modelling [10]. To address these limitations, we propose a unified, real-time framework that fuses visual scene context with detailed pose dynamics. This integration supports more accurate and interpretable predictions of penalty kick direction across varied match scenarios.

# 3 Methodology

This study presents a multi-modal deep learning approach to predict the final ball placement in football penalty kicks. The methodology comprises several stages: (1) constructing a curated dataset from real-world match footage; (2) developing object detection and pose estimation models to extract input features; (3) training a hybrid CNN–LSTM architecture; and (4) evaluating its performance under various parameter configurations.

## 3.1 Data set creation

**Penalty kick event dataset**
A total of 154 match highlight videos were collected from broadcasting platforms and publicly available datasets. These videos featured key match moments—including goals, fouls, and penalty kicks—from both international fixtures and top-tier club competitions. In addition, 12 full match recordings were sourced from online sports archives. From this collection, individual penalty-kick events were manually identified and extracted to isolate only the relevant action sequences involving the penalty shooter. For each event, the final position of the ball—whether it entered the left, middle, or right side of the goal, as viewed from the goalkeeper's perspective—was manually annotated. In total, 755 distinct penalty-kick scenarios were compiled.

The extracted data were further processed to align with the input requirements of the proposed neural network model. Each video clip was segmented into a fixed-length sequence, beginning from the moment the referee signalled the penalty kick and ending just before the player made contact with the ball. The endpoint of each sequence was determined using a distance-based threshold between the kicker's foot and the ball. To support this process, a customised object detection model was developed to accurately identify and track key elements within each frame.

**Object detection model (YOLOv8) training dataset**
A dedicated dataset was prepared to support the training of a custom object detection model designed to identify key elements involved in penalty kick scenarios. Approximately 4,000 RGB frames were extracted from the collected video clips. Each frame was manually annotated with bounding boxes corresponding to four object classes: penalty shooter, goalkeeper, net (the goal), and ball. To improve model robustness and generalisation, data augmentation techniques were applied, including rotation, blurring, scaling, shearing, and adjustments to brightness and saturation. This process expanded the dataset to 6,300 annotated frames. The final dataset was split into three subsets: 70% for training and 15% each for validation and testing.

**Penalty kick approach sequence dataset (Model Input 01)**
The object detection results—the positions of the goal shooter, goalkeeper, ball, and net—were used to define precise frame sequences. Each sequence began when the referee signalled the penalty kick and ended just before the shooter made contact with the ball. The endpoint was determined algorithmically by tracking the distance between the kicker's foot (estimated via pose keypoints) and the ball. A distance-based threshold was applied to ensure the sequence captured only the preparatory motion leading up to the shot.

However, since object sizes in the frame varied depending on camera angle and zoom level, a fixed pixel-based threshold did not generalise well across all clips. For instance, players appearing smaller in the frame resulted in proportionally smaller measured distances, leading to inconsistent endpoint detection. To resolve this, the distance between the starting position of the ball and the midpoint of the net—two reliably detected objects—was used as the reference, as their physical separation remains constant during the early phase of the penalty kick. The foot-to-ball distance threshold was then expressed as a ratio relative to this fixed reference, allowing for consistent sequence segmentation across varying perspectives and resolutions (Fig. 1).

To evaluate how the endpoint threshold influences the model prediction performance, 03 separate sets of video sequences were created using different normalised threshold ratios. The trajectory prediction model was trained independently on each dataset.

**Pose keypoint dataset of the penalty shooter (Model Input 02)**

The penalty shooter was identified using the object detection model, and their keypoints were extracted across the refined video segment using the YOLOv8-Pose algorithm—a robust single-stage, multi-person keypoint detector. YOLOv8-Pose accurately detects 17 standard body keypoints per frame (Fig. 1), corresponding to key anatomical landmarks such as the ankles, knees, hips, shoulders, elbows, wrists, neck, and head. This structured pose information was used as input to the skeletal feature branch of the proposed model.

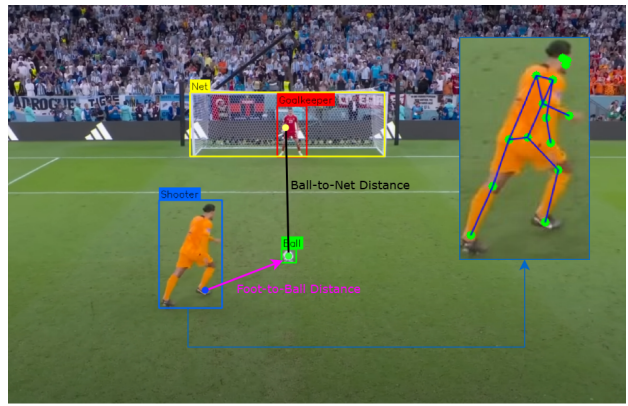

**Fig. 1.** Object detection results on a video frame with lines indicating distances from the ball to the net and to the shooter's nearest foot. A zoomed-in view displays the extracted pose keypoints of the shooter.

**Model input sequence finalisation**

The proposed neural network model required fixed-length input sequences of 8 frames, sampled from each video segment and its corresponding keypoint sequence.

To ensure that the model captured meaningful motion patterns, frames were not selected from the beginning of the sequence, as early frames often contained idle waiting periods and lacked informative biomechanical cues. Instead, frames were sampled uniformly across the entire duration of each video segment to provide a more representative view of the player's preparatory movement. For example, in a sequence containing 100 frames, 8 frames were selected at regular intervals (1, 15, 29, 43, 57, 71, 85, 99), enabling uniform temporal coverage. If the pose estimation model failed to detect keypoints in a selected frame with an average confidence score above 0.6, that frame was replaced by the nearest valid frame within the sequence.

The final dataset consisted of 755 samples. Each sample included two inputs: a sequence of 8 RGB frames and a keypoint tensor capturing the (x, y) coordinates of 17 body keypoints across the same 8 frames. The dataset was then divided into three subsets: 70% for training, 15% each for validation and testing.

### 3.2 Neural network architecture

The model processes two synchronised input streams extracted from each penalty kick sequence: (1) a sequence of RGB video frames with shape (B, 8, 224, 224, 3), where B is the batch size and 8 represents the temporal dimension; and (2) a corresponding sequence of 2D body keypoints with shape (B, 8, 17, 2), representing the (x, y) coordinates of 17 anatomical joints per frame. These inputs are passed through a custom hybrid deep learning architecture implemented using the TensorFlow Functional API. The architecture consists of four main components: the Spatial Feature Branch, Skeletal Feature Branch, Pose-Guided Spatial Attention Module, and the Late Fusion and Classification Head (Fig. 2). The design enables parallel processing of visual and pose information, with attention mechanisms guiding the model to focus on task-relevant spatial and temporal features. The outputs from both streams are then integrated through a late fusion module, which generates the final prediction across three goal direction categories.

### 1. Spatial feature branch – Visual data processing

The spatial stream is responsible for analysing the visual dynamics of the penalty kick from RGB video frames. Each frame is individually processed using a MobileNetV2 convolutional neural network wrapped in a time-distributed layer to preserve temporal consistency. This produces spatial feature maps for each frame. These features are then refined using a pose-guided attention mechanism that highlights areas of interest based on the kicker's body posture. The refined features undergo global average pooling to produce concise descriptors, which are then passed through a multi-head self-attention layer to capture temporal dependencies. An LSTM layer further processes this sequence to summarise the visual stream into a single vector representation.

**Fig. 2.** Proposed multi-modal architecture: visual and pose inputs are processed in parallel with spatial and temporal attention, then fused for final kick direction prediction.

2. **Skeletal feature branch – 2D body keypoints processing**

   The skeletal stream complements the visual analysis by modelling the biomechanics of the penalty shooter using pose keypoints. In each frame, the 2D keypoints are flattened into vectors, forming a temporal sequence that represents the player's motion pattern. A multi-head attention layer is then applied to capture key temporal dependencies in the movement, such as foot orientation, leg swing, or hip rotation. The attention-weighted sequence is then processed by an LSTM network, which outputs a summary vector that encodes the overall pose dynamics across the sequence.

3. **Pose-guided spatial attention module**

   This module bridges the spatial and skeletal streams by allowing the pose information to influence the visual attention mechanism. It generates a dynamic attention map for each frame by transforming the pose features and combining them with the visual feature maps using convolutional layers. These attention maps act as spatial filters that instruct the visual stream where to focus more precisely, often highlighting regions such as the kicking foot, the ball, the plant foot area and body orientation. By applying this pose-informed focus, the visual features become more task-specific and informative for prediction.

4. **Late fusion and classification head**

   In the final stage, the outputs from the spatial and skeletal streams are combined to produce the final prediction. The two summary vectors—one from each stream—are concatenated and passed through a fusion block comprising batch normalisation, dense layers, and dropout for regularisation. This structure enables the network to learn meaningful interactions between visual and pose-based information. The final output is generated through a softmax layer, yielding the probability distribution over the three goal zones: left, middle, and right.

The final model comprised 57 million trainable parameters while balancing model complexity with performance on the penalty kick direction prediction task.

### 3.3     Model training

The model was trained end-to-end using the Adam optimiser with a learning rate of 0.001. The Categorical Crossentropy loss function was used to support the multi-class classification task of predicting shot direction (left, middle, or right). To promote generalisation and prevent overfitting, the training data was shuffled at the start of each epoch. A batch size of 32 was chosen to balance computational efficiency with gradient stability. The model was trained for up to 100 epochs, with early stopping applied based on validation loss. Training was terminated if no improvement was observed over 10 consecutive epochs, and the model checkpoint with the lowest validation loss was saved for final evaluation.

# 4 Results and Evaluation

## 4.1 Object detection model

The custom-trained YOLOv8 object detection model achieved strong performance on the test set, with an mAP@0.5 of 0.935, a precision of 0.984, and a recall of 0.916. These results indicate reliable detection of the penalty shooter, ball, goalkeeper, and net, providing accurate inputs for tracking and subsequent analysis.

## 4.2 Penalty kick direction prediction model

To assess the impact of temporal sequence length, models were trained using three different normalised foot-to-ball distance thresholds. These thresholds were defined as ratios relative to the fixed distance between the kicker and the net, ensuring consistency across varying camera perspectives. Higher threshold values captured only the early stages of preparation, while lower thresholds included the full approach phase, potentially providing richer motion cues for trajectory prediction. Table 1 summarises the number of training iterations and the corresponding prediction accuracy on the test set for each threshold configuration. All models initially demonstrated effective learning; however, signs of overfitting were observed in each scenario, and training was halted via early stopping before reaching the maximum of 100 epochs. The final model checkpoints were then evaluated on a held-out test set. Fig. 3 presents the confusion matrices for each threshold, offering a more detailed view of the model's classification performance on the test set, which consisted of 113 samples.

**Table 1.** Model performance across varying foot-to-ball distance thresholds

| Distance threshold (Normalised ratio) | Training iterations | Testing accuracy (%) |
|---|---|---|
| 0.15 | 77 | 89.38 |
| 0.25 | 72 | 76.11 |
| 0.35 | 75 | 60.18 |

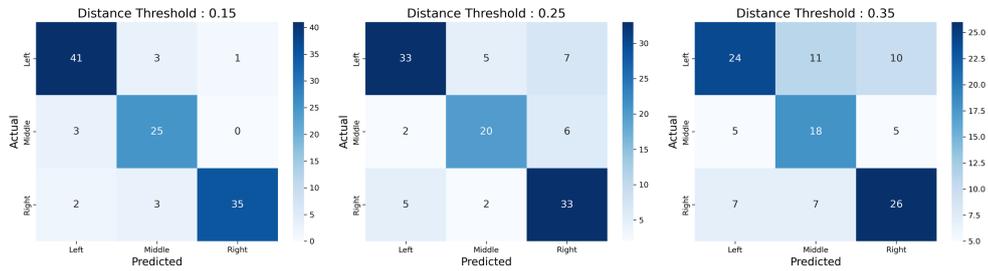

**Fig. 3.** Confusion matrices for different foot-to-ball distance threshold configurations

To evaluate the contribution of each model component, ablation studies were conducted using the dataset generated with a distance threshold of 0.15. Four model variants were compared:

**1.** Visual-only model: Used only the RGB frame sequence as input (spatial branch).
**2.** Pose-only model: Used only the player's keypoints as input (skeletal branch).
**3.** Dual-branch without pose-guided attention: Combined both input streams but excluded the pose-guided attention module.
**4.** Final proposed model: Full architecture with both spatial and pose branches, along with the pose-guided attention.

**Table 2.** Ablation study comparing test accuracy across different model configurations

| Model configuration | Test accuracy (%) |
|---|---|
| Visual-only (spatial feature branch) | 75.22 |
| Pose-only (skeletal feature branch) | 68.14 |
| Dual-branch (no pose-guided attention) | 82.30 |
| Final proposed model (with attention) | 89.38 |

The results, summarised in Table 2, show that while each individual branch contributes useful information, the visual-only configuration achieved a test accuracy of 75%, whereas the pose-only stream, based solely on body keypoints, reached 68%, indicating that visual cues alone are more informative than pose features when used in isolation.

However, when both streams were fused in the dual-branch model without attention, the accuracy increased to 82%, confirming that the integration of visual and biomechanical features leads to more robust predictions. The final proposed model, which incorporates pose-guided spatial attention, further improved accuracy to 89%. This highlights the added value of the attention mechanism in enabling the network to selectively focus on task-relevant areas of the input. The attention map generated by the pose-guided mechanism (Fig.4) visually confirms the model's ability to concentrate on the most relevant areas of the scene during a penalty kick.

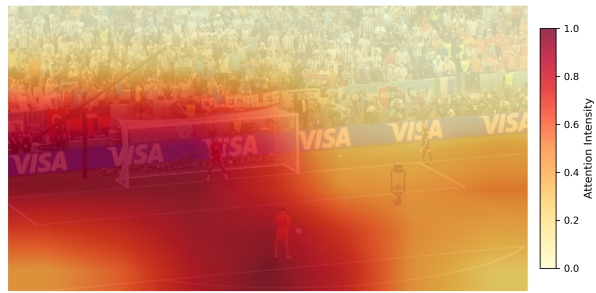

**Fig. 4.** Pose-guided spatial attention visualised as a heat-map overlay for a representative penalty-kick frame

### 4.3    Inference Pipeline

To evaluate the model's inference performance, the trained model was deployed on an NVIDIA RTX 4080 GPU and tested using real-world penalty kick footage. The inference framework, illustrated in Fig. 5, processed a continuous video stream. It dynamically trimmed the footage based on a user-defined distance threshold by monitoring the relative distance between the kicker's foot and the ball, and then passed the extracted input sequence to the model to generate the prediction.

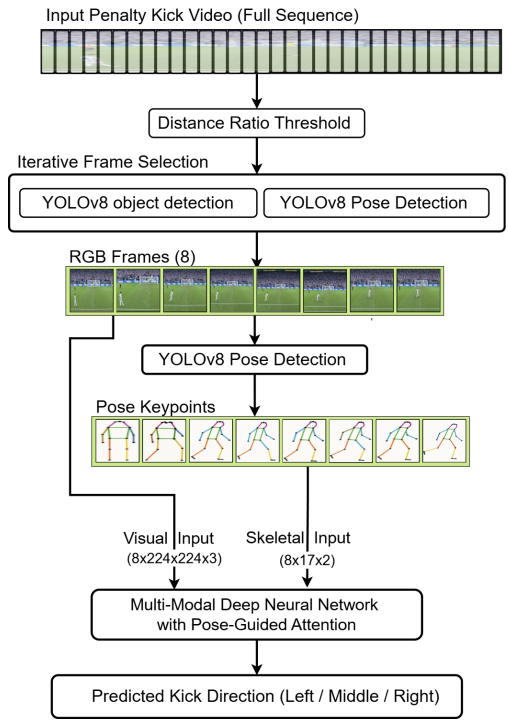

**Fig. 5.** End-to-end inference pipeline for penalty kick direction prediction using multi-modal inputs

To enable this process, a custom-trained YOLOv8 object detection model was used to detect key elements in each video frame, including the penalty taker, ball, goalkeeper, and goal net. Based on these detections, the distance from the ball to the net was calculated. At the same time, pose keypoints were extracted using YOLOv8-Pose to identify the kicker's foot closest to the ball. The distance from this foot to the ball was measured, and the ratio of foot-to-ball distance to ball-to-net distance was continuously monitored. When this ratio reached the user-defined threshold, the corresponding video segment was extracted. From this segment, 8 frames were selected to represent the player's final approach. For each frame, the 2D body keypoints of the penalty shooter were obtained. If a frame lacked reliable keypoints, it was replaced with the nearest valid frame to ensure a complete and consistent set of 8 RGB frames and their corresponding pose data.

These synchronised inputs—video frames and player keypoints—were then fed into the trained multi-modal neural network. The model produced a probability distribution over the three goal zones: left, middle, and right, as shown in Fig. 6. The entire inference process was completed in just 22 milliseconds.

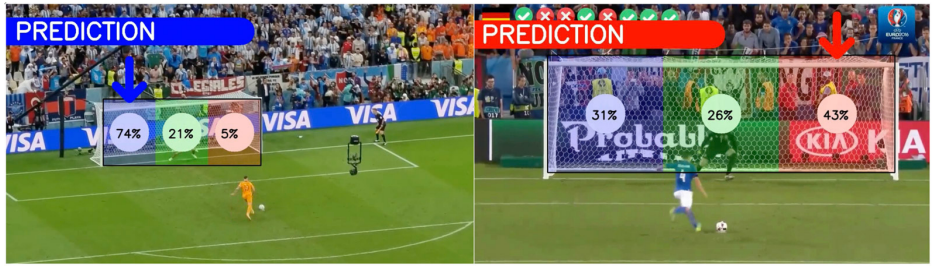

**Fig. 6.** Output from the trained model showing the probability distribution across the three goal zones

## 5    Discussion

The testing results highlight the effectiveness of the proposed multi-modal deep learning architecture for penalty kick direction prediction. By integrating visual context and player biomechanics through RGB frames and pose keypoints, the model captures detailed spatio-temporal patterns that are often overlooked by traditional approaches. The ablation studies confirm this advantage: models trained on only one modality—either visual or skeletal—achieved significantly lower accuracy compared to the combined model. This demonstrates that the spatial scene elements and player motion dynamics contribute complementary information essential for accurate prediction.

The inclusion of attention mechanisms further enhances performance. The pose-guided spatial attention and temporal self-attention modules allow the model to selectively focus on the most informative regions and moments during the penalty kick. As shown in Fig. 4, the pose-guided attention mechanism does not merely focus on the kicking foot but also highlights surrounding context such as the ball–foot interaction zone, the plant-foot position, upper-body orientation, and the relative positioning of the goalkeeper and goal. This context-aware attention mimics how human observers analyse action sequences and enables the model to effectively exploit the interplay between biomechanics and scene geometry.

With an inference time of 22 milliseconds per segment, the model is well-suited for real-time applications. It maintains a modest parameter count of 57 million, which is relatively low compared to similar dual-branch architectures with attention mechanisms. This contributes to its overall lightweight design, ensuring a balance between accuracy and computational efficiency, making it practical for goalkeeper training, match analysis, and tactical decision-making.

This work relies on a distance-based thresholding method to define the endpoint of each input sequence, capturing the moment just before the player strikes the ball. Unlike fixed-duration windows, which assume uniform camera perspectives and player speeds, the proposed approach adapts dynamically by monitoring the relative distance between the foot and the ball, normalised by the kicker–goal distance. This ensures consistent capture of the most informative preparatory phase—when body posture, foot angle, and momentum signal shot direction, regardless of run-up style or video perspective.

To further evaluate the impact of input sequence segmentation on performance, the model was tested with three different distance thresholds—0.15, 0.25, and 0.35—each representing how close the kicker's foot must be to the ball. The results show a clear trend: the smaller the threshold (the closer to the kick), the higher the prediction accuracy, with the 0.15 ratio achieving the best test accuracy of 89%. This setting captures the most critical biomechanical indicators, such as trunk rotation, kicking foot angle, and upper-body posture, all of which are strong cues for shot direction. At larger thresholds (0.25 and 0.35), the input sequence includes more of the approach phase, which often contains generic motion patterns or idle movement. As a result, the model receives less directly relevant information for the prediction task, leading to reduced test accuracy—76% and 60%, respectively.

However, it is noteworthy that even with these lower accuracies, the model still performs significantly above chance, indicating that useful predictive cues exist earlier in the run-up as well. Movements such as the approach angle, pace, posture, and weight distribution begin forming early and provide initial indicators of the kicker's intent, even if they are less explicit than the cues near the final strike. This insight proves valuable in real-world applications, where an early but approximate prediction can still influence strategic decision-making. For instance, even a moderately confident early prediction generated by our model, based solely on the initial phase of the run-up, can help a goalkeeper begin shifting their weight or adjusting their stance, thereby gaining critical milliseconds in reaction time.

Since the model analyses both pose dynamics and visual context over time, it can learn player-specific behavioural cues such as approach angle, trunk orientation, and foot positioning that tend to correlate with shot direction. These cues, even when captured early in the sequence, offer useful signals. Goalkeepers can be trained using these early predictions to understand not only where a player is likely to shoot, but also when to commit to a movement. For instance, if the model consistently predicts "right" based on certain shoulder alignments or stride patterns two steps before the kick, training can focus on timing dives as soon as these patterns emerge. This enables the development of player-specific anticipation strategies, helping goalkeepers learn to respond differently depending on the unique habits of each player.

On the offensive side, coaches can use the model's attention visualisations to highlight which early-phase movements are most predictive of direction. This can guide players in refining or disguising their approach—for example, adjusting foot placement or body posture to make their intended direction less detectable. In both cases, the model's ability to fuse spatial, temporal, and biomechanical data provides a rich foundation for strategic training grounded in real-world match behaviour.

# 6      Limitations

While the proposed multi-modal deep learning architecture demonstrates strong performance in predicting penalty kick direction, several limitations remain. The model's effectiveness depends heavily on clear visibility of the goalpost, ball, goalkeeper, and especially the shooter. Variability in camera angles, occlusions, or zoom levels can hinder detection and pose estimation accuracy, affecting performance in real-world broadcast conditions. Additionally, the model currently classifies shots into only three broad categories—left, middle, or right—limiting its tactical utility. Future work could explore finer-grained predictions, such as shot height, ball spin, or impact location within the goal, possibly supported by ball tracking or segmentation models.

Another limitation concerns domain generalisability. The model was trained on curated video segments, and its performance across varied leagues, player styles, or live broadcast conditions remains uncertain. This study establishes an initial step by creating a dedicated penalty kick dataset; however, expanding it with more diverse examples and real-time match scenarios will be necessary to improve generalisability and support broader applicability.

# 7      Conclusion

This study introduced a multi-modal deep learning framework for predicting penalty kick direction by combining visual context with pose-based biomechanical information. The dual-branch architecture integrates a MobileNetV2-based CNN for visual processing and an LSTM with attention mechanisms for temporal modelling. A pose-guided spatial attention module enhances the model's ability to focus on task-relevant regions, while a distance-based thresholding strategy ensures consistent input segmentation prior to ball contact. The model achieved 89% accuracy with a 22 ms inference time, offering a lightweight and efficient solution suitable for real-time deployment, with promising applications in goalkeeper training, match strategy development, and broader sports analytics.

**Disclosure of Interests.** The authors have no competing interests to declare that are relevant to the content of this article.

# References

1. Zheng, F., Al-Hamid, D.Z., Chong, P.H.J., Yang, C., Li, X.J.: A Review of Computer Vision Technology for Football Videos. Information 16, 355 (2025)
2. Sharma, V., Gupta, M., Pandey, A.K., Mishra, D., Kumar, A.: A Review of Deep Learning-based Human Activity Recognition on Benchmark Video Datasets. Applied Artificial Intelligence 36, (2022)
3. Pinheiro, G.D.S., Jin, X., Costa, V.T.D., Lames, M.: Body Pose Estimation Integrated With Notational Analysis: A New Approach to Analyze Penalty Kicks Strategy in Elite Football. Frontiers in Sports and Active Living 4, (2022)

4.  Host, K., Ivašić-Kos, M.: An overview of Human Action Recognition in sports based on Computer Vision. Heliyon 8, e09633 (2022)
5.  Cioppa, A., Deliège, A., Giancola, S., Ghanem, B., Droogenbroeck, M., Gade, R., Moeslund, T.: A Context-Aware Loss Function for Action Spotting in Soccer Videos (2020)
6.  Qiu, Z., Yao, T., Mei, T.: Learning Spatio-Temporal Representation with Pseudo-3D Residual Networks. (2017)
7.  Honda, Y., Kawakami, R., Yoshihashi, R., Kato, K., Naemura, T.: Pass Receiver Prediction in Soccer using Video and Players' Trajectories. In: 2022 IEEE/CVF Conference on Computer Vision and Pattern Recognition Workshops (CVPRW), pp. 3502-3511.
8.  Chakraborty, D., Kaushik, M.M., Akash, S.K., Zishan, M.S.R., Mahmud, M.S.: Deep Learning-Based Prediction of Football Players' Performance During Penalty Shootout. In: 2023 26th International Conference on Computer and Information Technology (ICCIT), pp. 1-6.
9.  Mauricio Salazar, J.A., Alatrista-Salas, H.: Football Penalty Kick Prediction Model Based on Kicker's Pose Estimation. In: ACM International Conference Proceeding Series, pp. 196-203.
10. Secco Faquin, B., Teixeira, L.A., Coelho Candido, C.R., Boari Coelho, D., Bayeux Dascal, J., Alves Okazaki, V.H.: Prediction of ball direction in soccer penalty through kinematic analysis of the kicker. J Sports Sci 41, 668-676 (2023)